\documentclass[runningheads]{llncs}
\usepackage{graphicx}
\usepackage{multirow}

\usepackage{multirow}
\usepackage{tabu}
\usepackage{bbm}
\usepackage{diagbox}
\usepackage[english]{babel}
\usepackage{comment}
\usepackage{array}
\usepackage{amstext}
\usepackage{makecell}
\usepackage{caption}
\usepackage{tablefootnote}
\usepackage{babel}
\usepackage{booktabs}

\usepackage[ruled,vlined]{algorithm2e}
\newcommand{\nosemic}{\renewcommand{\@endalgocfline}{\relax}}
\newcommand{\dosemic}{\renewcommand{\@endalgocfline}{\algocf@endline}}
\usepackage{chngcntr}
\usepackage{placeins}
\usepackage{amsfonts}       
\usepackage{nicefrac} 
\usepackage{amsmath,amssymb} 
\usepackage{amstext}

\SetKwInput{KwInput}{Input}                
\SetKwInput{KwOutput}{Output}              
\usepackage[pagebackref=true,breaklinks=true,letterpaper=true,colorlinks,bookmarks=false]{hyperref}
\newlength\mylen

\newcommand{\ie}{\textit{i.e.}}
\newcommand{\eg}{\textit{e.g.}}
\newcommand{\alignshift}{\emph{AlignShift}}

\begin{document}
\title{\alignshift: Bridging the Gap of Imaging Thickness in 3D Anisotropic Volumes}
\titlerunning{Bridging the Imaging Thickness Gap with \alignshift}

\author{Jiancheng Yang\inst{1,2,3,}\thanks{These authors have contributed equally: Jiancheng Yang and Yi He.}  \and Yi He\inst{3,\star} \and Xiaoyang Huang\inst{1,2} \and Jingwei Xu\inst{1,2} \and \\Xiaodan Ye\inst{4} \and Guangyu Tao\inst{4} \and Bingbing Ni\inst{1,2,5,}\thanks{Corresponding author: Bingbing Ni (nibingbing@\{sjtu.edu.cn,hisilicon.com\}).}}

\authorrunning{J. Yang et al.}
%
\institute{Shanghai Jiao Tong University, Shanghai, China\\
	\and MoE Key Lab of Artificial Intelligence, AI Institute, Shanghai Jiao Tong University\\
	\and Dianei Technology, Shanghai, China\\
	\and Shanghai Chest Hospital, Shanghai Jiao Tong University School of Medicine
	\and Huawei Hisilicon, Shanghai, China	\\
	\email{jekyll4168@sjtu.edu.cn, yi.he@dianei-ai.com}	
}

\maketitle              
\begin{abstract}

This paper addresses a fundamental challenge in 3D medical image processing: how to deal with imaging thickness. For anisotropic medical volumes, there is a significant performance gap between thin-slice (mostly 1mm) and thick-slice (mostly 5mm) volumes. Prior arts tend to use 3D approaches for the thin-slice and 2D approaches for the thick-slice, respectively. We aim at a unified approach for both thin- and thick-slice medical volumes. Inspired by recent advances in video analysis, we propose \alignshift, a novel parameter-free operator to convert theoretically any 2D pretrained network into thickness-aware 3D network. Remarkably, the converted networks behave like 3D for the thin-slice, nevertheless degenerate to 2D for the thick-slice adaptively. The unified thickness-aware representation learning is achieved by shifting and fusing aligned ``virtual slices'' as per the input imaging thickness. Extensive experiments on public large-scale DeepLesion benchmark, consisting of 32K lesions for universal lesion detection, validate the effectiveness of our method, which outperforms previous \emph{state of the art} by considerable margins without whistles and bells. More importantly, to our knowledge, this is the first method that bridges the performance gap between thin- and thick-slice volumes by a unified framework. To improve research reproducibility, our code in PyTorch is open source at \url{https://github.com/M3DV/AlignShift}.

\end{abstract}

\section{Introduction} \label{sec:intro}

\begin{figure}[tb]
	\includegraphics[width=\linewidth]{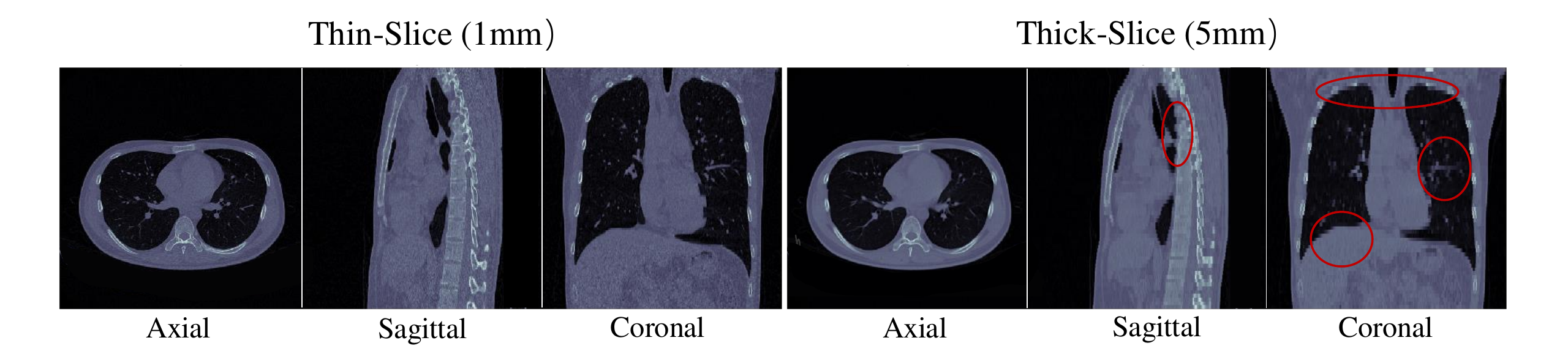}
	\caption{Illustration of spatially normalized thin- and thick-slice computed tomography (CT) scans of a same subject. Data is from our custom dataset. Note that even though there is no significant difference between the axial slices, artifacts (the highlights on the plots) are not neglectable for the sagittal and coronal views in the thick-slice data, which results in domain shift for 3D approaches.}
	\label{fig:thin_thick_cts}
\end{figure}

Deep learning has been dominating medical image analysis research in a wide range of tasks (\eg, classification \cite{gulshan2016development,zhao20183d,zhao2019toward}, segmentation \cite{isensee2018nnu,tang2019clinically}, detection \cite{tang2019nodulenet,yang2020relational}, registration \cite{balakrishnan2019voxelmorph,dalca2018unsupervised}). However, deployment of the medical image AI systems is still challenging due to numerous difficulties, \eg, open set scenarios \cite{Scheirer2013TowardOS}, calibration and uncertainty quantification \cite{Guo2017OnCO,huang2019evaluating} in real-world distribution, label ambiguity in clinical annotations \cite{kohl2018probabilistic,yang2019probabilistic}. In this study, we focus on a fundamental issue in 3D medical image analysis: how to deal with the imaging thickness, which denotes the physical distance between axial slices. In practice, there exist both thin-slice (mostly 1mm) and thick-slice (mostly 5mm) for a same task, \eg, lesion detection \cite{yan2018deep}, organ and tumor segmentation \cite{Simpson2019ALA}. Standard procedure treats this issue as pre-processing; spatial normalization is commonly applied to normalize the dataset into a same reference thickness (\eg, 2mm). However, the spatial normalization may amplify unwanted noises in medical images \cite{Glocker2019MachineLW}. Fig.~\ref{fig:thin_thick_cts} depict spatially normalized thin- and thick-slice computed tomography (CT) scans of a same subject. As illustrated, spatial normalization introduces significant artifacts to the thick-slice (note the sagittal and coronal views). If the spatially normalized thin- and thick-slice volumes are processed by a same CNN with standard convolutions, it will lead to domain shift. We conjecture that it is the reason why 3D approaches are preferred for thin-slice volumes, while 2D approaches tend to be superior for thick-slice/anisotropic volumes \cite{isensee2018nnu}. Spatial normalization for thick-slice data leads to larger information loss compared to that for thin-slice data. For this reason, we challenge the spatial normalization as a standard pre-processing procedure for 3D medical image processing. 

To address the thickness issue, we propose a novel parameter-free operator, \alignshift, to convert theoretically any 2D pretrained network into thickness-aware 3D network.
The proposed \alignshift~operator is inspired by Temporal Shift Module (TSM) in video analysis \cite{Lin2019TSMTS}, which enables temporal (or 3D) information fusion by shifting adjacent slices (details in Sec. \ref{sec:tsm}). Notably, TSM enables 2D-to-3D transfer learning, \ie, pretrained 3D networks on 2D datasets, which is also highly related to previous studies~\cite{liu20183d,yang2019reinventing}. Although superior to 2.5D approaches \cite{Yan2019MULANMU,zlocha2019improving,li2019mvp}, TSM does not bridge the performance gap between thin- and thick-slice volumes (see Sec. \ref{sec:performance-thin-thick}). As a comparison, the \alignshift~operator shifts and fuses aligned ``virtual slices'' as per the input imaging thickness, which results in unified thickness-aware representation learning (details in Sec. \ref{sec:alignshift}). Remarkably, the \alignshift-converted networks adaptively behave like 3D for the thin-slice, and like 2D for the thick-slice.

We validate the effectiveness of the proposed method on large-scale DeepLesion benchmark \cite{yan2018deep}, a universal lesion detection dataset with 3D inputs and key-slice annotations of 32K lesions. Without whistles and bells, the proposed methods outperform previous \emph{state of the art} \cite{Yan2019MULANMU,zlocha2019improving,li2019mvp} by considerable margins. More importantly, our method closes the performance gap between thin- and thick-slice volumes compared to both 2.5D and TSM approaches; to our knowledge, we are the first to achieve this by a unified framework.

\section{Methods}

\begin{figure}[tb]
	\includegraphics[width=\textwidth]{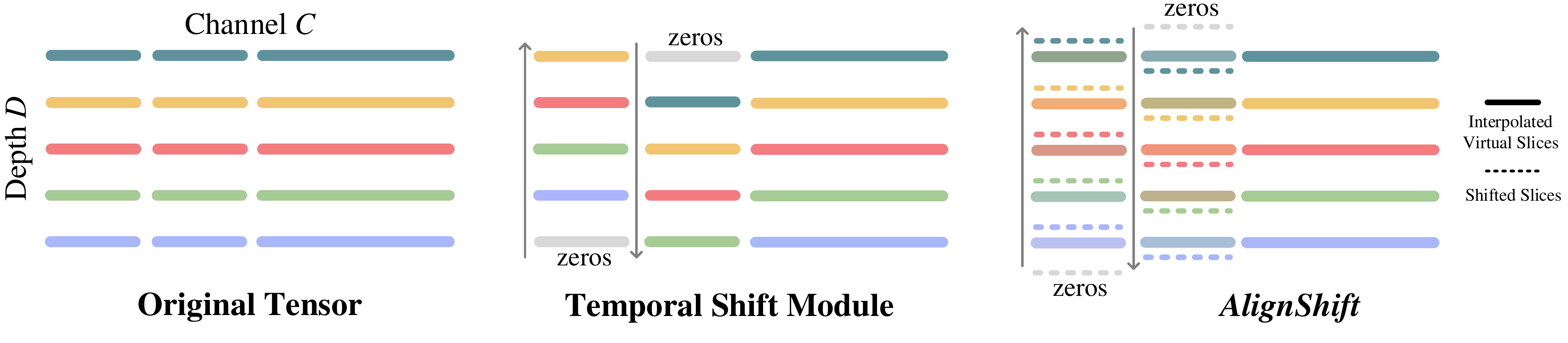}
	\caption{Illustration of Temporal Shift Module (TSM) \cite{Lin2019TSMTS} and the proposed \alignshift. \textbf{Left}: A 3D tensor ($C\times D\times H \times W$) with channel $C$, depth $D$, height $H$, width $W$, $H$ and $W$ are not depicted for simplicity. \textbf{Middle}: Temporal Shift Module (TSM). The channels are split into three parts for shifting up, shifting down and keeping original. Border slices are padded with zeros. \textbf{Right}: \alignshift. Instead of shifting physical slices in TSM, we shift ``virtual'' slices (solid lines in different colors) in \alignshift. The virtual slices are interpolated from shifted slices (dash lines) by a reference thickness $r$.} \label{fig:alignshift}
\end{figure}

\subsection{Preliminary: Temporal Shift Module (TSM)}  \label{sec:tsm}
 Prior arts in 3D images processing utilize pure 2D network to leverage the 2D pretrained weights, while 3D randomly-initialized network is necessarily adopted to fuse the feature into 3D representations. 2.5D representation, \ie, several slices as channels for 2D networks, is insufficient to capture 3D contexts. It is thus meaningful to directly convert a 2D pretrained network into a 3D counterpart. To this end, we introduce Temporal Shift Module \cite{Lin2019TSMTS} (TSM) from the field of video understanding, which enables 2D-to-3D network conversion. To our knowledge, this paper is the first to introduce TSM into medical images with proven effectiveness on DeepLesion \cite{yan2018deep} benchmark. TSM tries to leverage data shift operation \cite{Wu2017ShiftAZ} to capture 3D semantics under 2D CNN framework, in which data shift indicates shifting data along a dimension by a certain number of slices. In practice, it is inserted as an operator before a $1\times K\times K$ 3D convolution with kernel size $K$. Given a 3D tensor ($C\times D\times H \times W$) with channel $C$, depth $D$, height $H$, width $W$, TSM shifts the slices in the depth dimension by $+1$ slice in one part of the channels, and by $-1$ slice in another part of the channels, while the rest part of channels remain static (see Figure \ref{fig:alignshift} Middle). Information between slices is fused by channel in this way. To some extent, TSM imitates 3D approaches by slices shifting. It is capable of processing 3D data in an efficient way. However, for medical images, TSM faces the issue of various imaging thickness, which is widespread in many medical image datasets such as DeepLesion~\cite{yan2018deep}. TSM itself does not deal with this issue. Hence all the volumes are supposed to be normalized to the same thickness via interpolation before fed into a TSM model, no matter how large the volume imaging thickness is. To those volumes with thick-slice, the interpolation process could cause large distortion of information, as the highlighted artifacts showed in Figure \ref{fig:thin_thick_cts}. According to the extensive experiments in Section \ref{sec:experiments}, TSM has good performance over thin-slice volumes, while it declines enormously for thick-slice volumes.

\subsection{\alignshift} \label{sec:alignshift}

\begin{algorithm}[tb]
	
	\caption{In-place \alignshift~for 3D volumes (zero padding)}
	\label{algo:alignshift}
	\small
	\KwInput{input 3D feature $\boldsymbol{X} \in \mathbb{R}^{C\times D\times H\times W}$, input actual thickness $s\in \mathbb{R}^+$, shift channels $C^{+},C^{-},~(C^{+}+C^{-}<C)$, reference thickness $r\in \mathbb{R}^+$.
	}
	\KwOutput{$\boldsymbol{X} \in \mathbb{R}^{C\times D\times H\times W}$ (in-place assignment).}
	
	\nl Compute align factor: $\alpha=r/s$;
	
	\nl Shift up: $\boldsymbol{X^{+}} = [\boldsymbol{X}[:C^{+}], \boldsymbol{0}] \in \mathbb{R}^{C^{+}\times (D + 1)\times H\times W}$;
	
	\nl Obtain virtual slices: $\boldsymbol{X}[:C^{+}] = \alpha\cdot X^{+}[:,1:] + (1-\alpha)\cdot X^{+}[:,:-1]$;
	
	\nl Shift down: $\boldsymbol{X^{-}} = [\boldsymbol{0},\boldsymbol{X}[C^{+}:(C^{+}+C^{-})]] \in \mathbb{R}^{C^{-}\times (D + 1)\times H\times W};$
	
	\nl Obtain virtual slices: $\boldsymbol{X}[C^{+}:(C^{+}+C^{-})] = \alpha\cdot X^{-}[:,:-1] + (1-\alpha)\cdot X^{-}[:,1:].$
	
\end{algorithm}

We believe that spatial normalization by interpolation induces the performance gap between thin- and thick-slices volumes (see results in Table \ref{tab:thin-thick}). Domain shift takes place when normalizing thick-slices volumes, which damages the performance. To address the thickness issue, we introduce virtual slices and propose {\alignshift} which enables adaptive data shift operation based on the given imaging thickness. \alignshift~avoids spatial normalization to thick-slice volumes by treating thin- and thick-slice volumes separately. Without loss of generality, we define thick volumes as volumes that have a thickness larger than a reference thickness $r$, and vice versa. For thin-slice volumes, we normalize the thickness to the reference thickness $r$ by interpolation as usual. For thick-slice volumes which could be easily skewed by interpolation, their original thicknesses $s$ are kept. Given a 3D feature tensor $\boldsymbol{X}$ of shape $C\times D\times H \times W$ with channel $C$, depth $D$, height $H$, width $W$ and physical thickness $s$ on the depth dimension, {\alignshift} shifts part of channels up (denoted as $C_+$), and another part of channels down (denoted as $C_-$) along the depth dimension, while the rest channels remain static. In order to maintain a consistent ``receptive field'' in  physical sense along the depth dimension, it shifts the data by a continuous step, whose step size (align factor) $\alpha=r/s$ depends on the reference thickness $r$ and the volume's actual thickness $s$. As illustrated in Figure \ref{fig:alignshift}, the shifted slice, called ``virtual slice'', is obtained by interpolation between the adjacent two slices. See Algorithm \ref{algo:alignshift} for the mathematical formulation. Compared to our method, TSM discretely shifts the data by one full slice. The data shift strategy of TSM results in an inconsistent ``receptive field'' along the depth dimension in convolution, given the non-unified thickness. Thereby the data shift operation of TSM is so-called ``unaligned'' under this situation. Contrarily, our method aligns the 3D features of various thickness, allowing the network to learn thickness-aware representations with the same kernels. It is guaranteed by \alignshift~that the physical distance between shifted and un-shifted channels is always consistent among volumes with different thickness. \alignshift~bridges the performance gap between thin-slice and thick-slick CTs theoretically and empirically.

In practice, \alignshift~is simple to use and implement. Similar to TSM, it is inserted as an operator before a $1\times K\times K$ 3D convolution. No additional spatial normalization is needed. The original thickness is sent to the network to allow adaptive data shift operation. Compared to TSM, only modest modification is needed to gain a great performance boost. Note that \alignshift~is able to capture 3D semantics like TSM, while it could degenerate to 2D for data with extremely large thickness, as a result of the align factor close to zero.

\alignshift~serves as a parameter-free operator enabling to convert theoretically any 2D pretrained network into thickness-aware 3D network. The conversion process is straight-forward. Table \ref{tab:2d-to-3d} lists how main operators in 2D CNNs are converted to the counterpart in 3D CNNs.

\subsection{3D Network for Universal Lesion Detection on Key Slices}

\begin{table}[tb]
	\centering
	\caption{Convert a pretrained 2D backbone into 3D. We use DenseNet-121 \cite{huang2017densely}; \alignshift~is only applied in the dense blocks. $K$ denotes the kernel size.} \label{tab:2d-to-3d}
	\begin{tabular*}{\hsize}{@{}@{\extracolsep{\fill}}lccc@{}}
		\toprule
		2D Backbone & Conv2D $K\times K$  & Pool2D $K\times K$ & Norm2D \\
		\midrule
		3D Backbone  & (\alignshift +)Conv3D $1\times K\times K$ & Pool3D $1\times K\times K$  & Norm3D \\
		\bottomrule
	\end{tabular*}
\end{table}

\begin{figure}[tb]
    \centering
	\includegraphics[width=\textwidth]{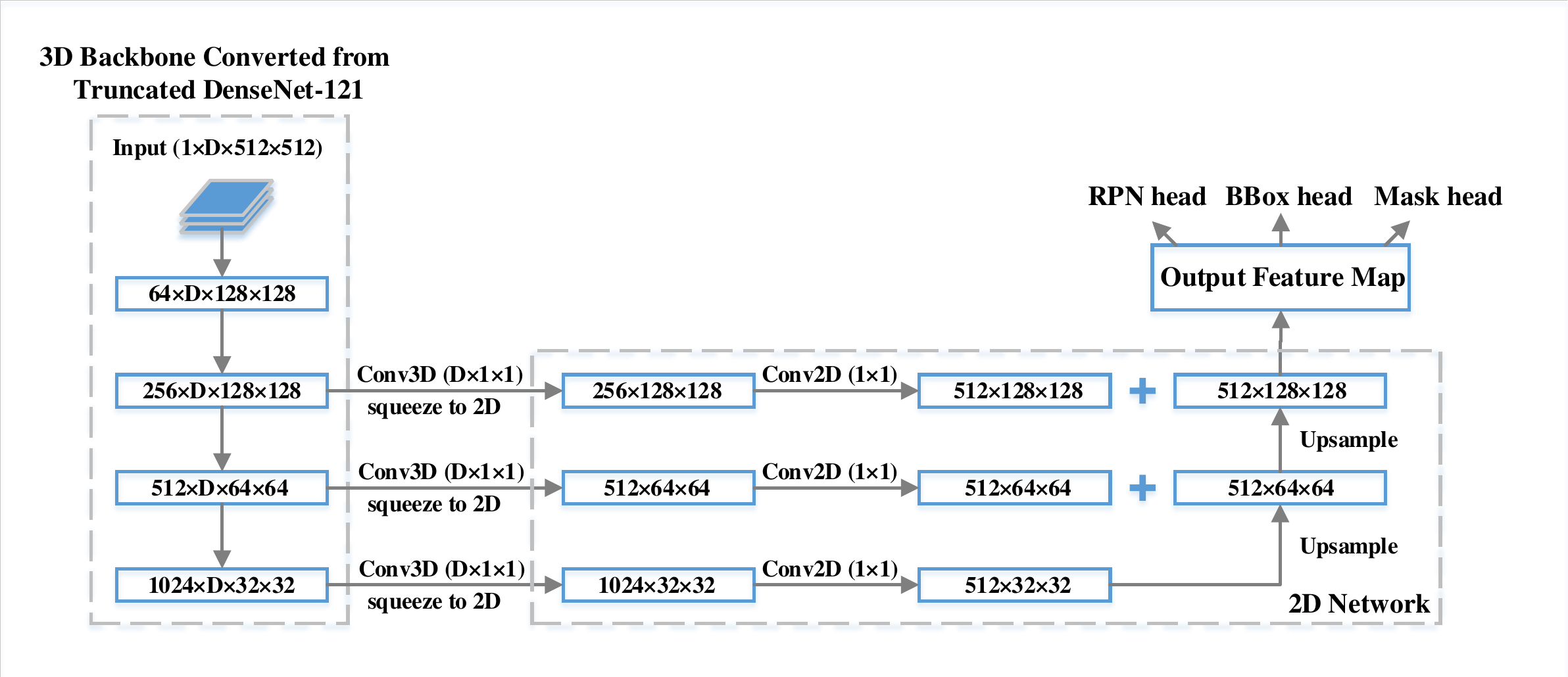}
	\caption{The 3D network for universal lesion detection on DeepLesion~\cite{yan2018deep}. The 3D backbone converted by a truncated DenseNet-121~\cite{huang2017densely,Yan2019MULANMU} processes a grey-scale 3D input of $D\times 512 \times 512$, where $D$ is the length of slices ($D=3,7$ in this study). 2D features of key slices are extracted, and then upsampled in a feature pyramid~\cite{Lin2016FeaturePN}. Detection is based on instance segmentation by Mask-RCNN~\cite{He2017MaskR}. Weak segmentation ``ground truth'' are generated from weak RECIST labels~\cite{zlocha2019improving}.} \label{fig:network}
\end{figure}

We experiment with the proposed method on DeepLesion benchmark \cite{yan2018deep}, which is a large-scale dataset on universal lesion detection. The inputs are 3D slices whereas only 2D key-slice annotations are available. We develop a 3D network with 2D detection heads, based on Mask RCNN \cite{He2017MaskR}. As illustrated in Figure \ref{fig:network}, the 3D backbone is converted from the truncated DenseNet-121 \cite{huang2017densely} with \alignshift, serving as a 3D feature encoder. All the $K\times K$ 2D convolutions in the dense blocks are converted to \alignshift $+ 1\times K\times K$ 3D convolutions. The encoder takes a grey-scale 3D tensor of $1\times D\times 512\times 512$ as input, where $D$ is the length of key slices ($D=3,7$ in this study), and extracts 3D features through three dense blocks. Each dense block increases the feature channels, downsamples the feature in the height and width dimension while maintaining the scale in the depth dimension. The feature output of each dense blocks is processed by a $D\times 1\times 1$ 3D convolution and then squeezed into a 2D shape. A 2D decoder combines these features under different resolutions and upsamples the features step by step. The final feature map is fed into RPN head, BBox head and Mask head for detection and instance segmentation supervised by weak RECIST labels.

We implement the Mask-RCNN with PyTorch \cite{Paszke2017AutomaticDI} and MMDetection~\cite{mmdetection}. As counterparts of the proposed \alignshift, we also implement 1) 2.5D Mask-RCNN, which stacks the slices as input channels for standard 2D networks, and 2) TSM-converted Mask-RCNN, which uses TSM instead of \alignshift. Note that the model sizes of the TSM models are strictly the same as the \alignshift models, and the 2.5 models are slightly smaller because of the Conv3D ($D\times 1\times 1$) layer.

\section{Experiments}
\label{sec:experiments}
\subsection{Dataset and Experiment Setting}

DeepLesion dataset \cite{yan2018deep} consists of 32,120 axial CT slices from 10,594 studies of unique patients. the thickness of the dataset is almost 1mm (48.66\%) and 5mm (50.26\%), which is appropriate to develop and evaluate the proposed method. There are 1 to 3 lesions in each slice, with totally 32,735 lesions from several organs, whose sizes vary from 0.21 to 342.5mm. RECIST diameter (critical for tumor staging and prognosis assessment \cite{yang2019development}) coordinates and bounding boxes were labeled on the key slices, with adjacent slices (above and below 30mm) provided as contextual information. We use GrabCut~\cite{rother2004grabcut} to generate weak segmentation ``ground truth'' from weak RECIST labels~\cite{cai2018accurate,zlocha2019improving}. Hounsfield units of the input are clipped into $[-1024,2050]$ and normalized. In \alignshift~experiments, $r=2mm$ is regarded as the reference thickness; we normalize the thickness into 2mm for thin-slice data with thickness$<=$2mm; for thick-slice data with thickness$>$2mm, we keep the original thickness, since the spatial normalization for thick-slice data leads to larger information loss compared to that for thin-slice data. For 2.5D and TSM counterparts, we adapt standard strategy to normalize the thickness of all images into 2mm. Data augmentation including horizontal flip, shift, rescaling and rotation is applied during training stage, no test-time augmentation (TTA) is applied. We resize each input slice to $512\times512$ before feeding into the networks. We use official data split (training/validation/test: 70\%/15\%/15\%); following prior studies~\cite{Yan2019MULANMU,zlocha2019improving,li2019mvp}, sensitivity at various false positives levels (\ie, FROC analysis) is evaluated on the test set.

\begin{table}[tb]
	
	\caption{Sensitivity (\%) at various false positives (FPs) per image of previous \emph{state-of-the-art} and the proposed methods, on the large-scale DeepLesion benchmark \cite{yan2018deep}. Note that MULAN \cite{Yan2019MULANMU} uses extra tag supervision and an addition Score Refinement Layer (SRL) with tag inputs; We report the performance of MULAN under 171-tag supervision as well as that without SRL. }\label{tab:deeplesion-performance}
	
	\centering
	\begin{tabular*}{\hsize}{@{}@{\extracolsep{\fill}}lcccccccc@{}}
		\toprule
		Methods& Slices & 0.5 & 1 & 2 & 4 & 8 & 16 & Avg.[0.5,1,2,4] \\
		\midrule
		3DCE \cite{yan20183d}~MICCAI'18& $\times 27$ & 62.48 & 73.37 &80.70 &85.65 &89.09 &91.06& 75.55\\
		ULDor \cite{tang2019uldor}~ISBI'19& $\times 1$ & 52.86 & 64.8 & 74.84 & 84.38 & 87.17 & 91.8 & 69.22\\
		V.Attn \cite{wang2019volumetric}~MICCAI'19& $\times 3$ & 69.10 & 77.90 & 83.80 & - & - & - &-\\
		Retina. \cite{zlocha2019improving}~MICCAI'19& $\times 3$ & 72.15 & 80.07 & 86.40 & 90.77 & 94.09 & 96.32 & 82.35\\
		MVP \cite{li2019mvp}~MICCAI'19 & $\times 3$ & 70.01 & 78.77 &84.71 &89.03 &- &- &80.63  \\
		MVP \cite{li2019mvp}~MICCAI'19 & $\times 9$ & 73.83 & 81.82 & 87.60 & 91.30 & - & - & 83.64 \\
		MULAN \cite{Yan2019MULANMU}~MICCAI'19& $\times 9$  & 76.12 & 83.69 & 88.76 & 92.30 & 94.71 & 95.64 & 85.22 \\
		w/o SRL \cite{Yan2019MULANMU}~MICCAI'19& $\times 9$  & -& -&-&-&-&- & 84.22 \\
		\midrule
		Ours 2.5D & $\times 3$  & 71.27 & 79.82 & 86.30 & 90.61 & 93.75 & 95.70 & 82.00 \\ 
		Ours 2.5D &  $\times 7$  & 72.66 & 81.45 & 87.07 & 90.98 & 93.40 & 95.30 & 83.04 \\ 
		Ours TSM&  $\times 3$ & 70.24 &	79.52 &	86.28 &	90.90 &	94.06 &	96.09 &	81.73\\ 
		Ours TSM&  $\times 7$  & 75.98 & 83.65 & 88.44 & 92.14 & 94.89 & 96.50 & 85.05 \\  
		Ours \alignshift & $\times 3$ & 72.90 &	80.74 &	87.15 &	91.92 &	94.85 &	96.48 &	83.18  \\ 
		Ours \alignshift &  $\times 7$   & \bf79.40 & \bf85.50 &\bf90.09 &\bf93.26 &\bf95.24 &\bf96.66 &\bf87.06\\ 
		\bottomrule
	\end{tabular*}
	
\end{table}

\subsection{Performance Compared with \emph{State of the Art}}
In Table \ref{tab:deeplesion-performance}, we depict sensitivity at various false positives per image (FPs), which shows that our proposed methods significantly outperform the previous \emph{state-of-the-art} MULAN \cite{Yan2019MULANMU}. Notably, it is achieved without additional information beyond the CT images such as tags from medical reports and demographic information. The performance of our 2.5D counterparts is comparable to Improved RetinaNet~\cite{zlocha2019improving} and MVP-Net~\cite{li2019mvp}. The proposed TSM-converted networks outperform these studies~\cite{zlocha2019improving,li2019mvp}, and even MULAN without tag supervision, which validates the superiority of pretrained 3D backbones over 2.5D. \alignshift~further boosts performance and surpasses our TSM and previous \emph{state-of-the-art} MULAN \cite{Yan2019MULANMU}. Due to memory constraints, we report the performance of maximum 7 slices, whereas more slices are expected with better performance.

\begin{table}[tb]
	\centering
	\caption{Detection performance analysis of our methods on all, thin-slice (mostly 1mm) and thick-slice (mostly 5mm) data. ``diff.'' denotes the average sensitivity (Avg.[0.5,1,2,4]) difference between thin/thick-slice data and that of all data.}
	
	\label{tab:thin-thick}
	\begin{tabular*}{\hsize}{@{}@{\extracolsep{\fill}}lccccccccc@{}}
		\toprule
		Methods & Thinkness & 0.5 & 1 & 2 & 4 & 8 & 16 & Avg.[0.5,1,2,4] & diff. \\
		\midrule
		\multirow{3}{*}{2.5D $\times 3$} & All & 71.27 & 79.82 & 86.30 & 90.61 & 93.75 & 95.70 & 82.00 &-\\
		
		& Thin & 72.78 & 80.65 & 87.21 & 90.94 & 93.97 & 95.86 & 82.89 & $+0.89$ \\ 
		& Thick   & 69.88 & 79.16 & 85.51 & 90.48 & 93.65 & 95.65 & 81.26 & $-0.74$ \\
		\multirow{3}{*}{2.5D $\times 7$} & All & 72.66 & 81.45 & 87.07 & 90.98 & 93.40 & 95.30 & 83.04 & -\\
		& Thin & 75.77 & 83.93 & 88.85 & 92.37 & 94.26 & 95.78 & 85.23 &$+2.19$  \\ 
		& Thick   & 69.76 & 78.96 & 85.75 & 90.03 & 92.67 & 94.99 & 81.13 &$-1.91$ \\
		\midrule 
		
		\multirow{3}{*}{TSM $\times 3$} & All & 70.24 &	79.52 &	86.28 &	90.90 &	94.06 &	96.09 &	81.73 & - \\
		& Thin  & 73.74 & 81.84 &87.54 &	92.21 &	94.92 &	96.72 &	83.83 &	$+2.10$\\ 
		& Thick & 67.03 &	77.37 &	84.98 &	89.95 &	93.16 &	95.44 &	79.83 &	$-1.90$ \\ 
		
		\multirow{3}{*}{TSM $\times 7$} & All& 75.98 & 83.65 & 88.44 & 92.14 & 94.89 & 96.50 & 85.05 & -\\
		& Thin  & 78.76 & 85.53 & 89.67 & 93.48 & 95.61 & 96.68 & 86.86 &$+1.81$ \\ 
		& Thick   &73.26 & 81.97 & 87.10 & 90.96 & 94.14 & 96.42 & 83.32 &$-1.73$ \\ 
		\midrule
		
		\multirow{3}{*}{\alignshift~$\times 3$} & All& 72.90 &	80.74 &	87.15 &	91.92 &	94.85 &	96.48 &	83.18 & -\\
		& Thin & 73.51 &	81.59 &	87.62 &	92.37 &	94.87 &	96.51 &	83.77 &	$+0.61$ \\ 
		& Thick & 72.85 &	80.18 &	87.10 &	91.94 &	94.91 &	96.54 &	83.02 &	$-0.16$ \\
		\multirow{3}{*}{\alignshift~$\times 7$} & All&79.40 &	85.50 &	90.09 &	93.26 &	95.24 &	96.66 &	87.06  & -\\		
		& Thin & 80.73 &	86.43 &	91.02 &	93.97 &	95.78 &	96.97 &	88.04 &	$+0.98$ \\ 
		& Thick &78.27 &	84.74 &	88.89 &	92.47 &	94.67 &	96.38 &	86.09 &	$-0.97$ \\ 
		\bottomrule
	\end{tabular*}
	
\end{table}

\subsection{Performance Analysis on Thin-Slice and Thick-Slice Data} \label{sec:performance-thin-thick}

To demonstrate the benefits of \alignshift~on bridging the performance gap of thin- and thick-slice data, we conducted a performance analysis of our 2.5D, TSM and \alignshift~models on the thin- and thick-slice data separately. As there is no open trained models available for prior \emph{state of the art}~\cite{Yan2019MULANMU,zlocha2019improving,li2019mvp}, we believe our 2.5D models represent the performance of these studies, considering these studies follow 2.5D fashion. As illustrated in Table \ref{tab:thin-thick}, there is a significant performance gap between thin- and thick-slice CTs for 2.5D and TSM approaches, which validates our argument (Sec. \ref{sec:intro}) that there is a significant domain shift if the thin- and thick-slice data are processed by a same CNN with standard convolutions. Note that the gap becomes larger when more slices are used, since the 3D information is less valuable for thick-slice data than the thin-slice. In comparison, by learning unified thickness-aware representation, the proposed \alignshift~ reduces the performance gap to neglectable for 3-slice setting; even for the 7-slice setting, the gap is closed compared to 2.5D and TSM approaches.

\section{Conclusion}

In this study, we challenge spatial normalization as a standard pre-processing approach to solve thickness issue in 3D medical images. Our experiment results indicate that both 2.5D and 3D (\ie, TSM in this study) approaches do not fundamentally address the domain shift issue introduced by spatial normalization, which results in a significant performance gap between thin- and thick-slice data. In this regard, we propose a novel parameter-free operator \alignshift, which enables us to convert theoretically any 2D pretrained network into thickness-aware 3D network. Extensive experiments on DeepLesion benchmark empirically validate that our methods bridge the gap between thin- and thick-slice data. Without whistles and bells, we establish a new \emph{state of the art} on DeepLesion, which surpasses prior arts by considerable margins.

\subsubsection{Acknowledgment.}
This work was supported by National Science Foundation of China (61976137, U1611461). This work was also supported by Interdisciplinary Program of Shanghai Jiao Tong University (YG2017QN661). Authors appreciate the Student Innovation Center of SJTU for providing GPUs.

\bibliographystyle{splncs04}

\end{document}